\documentclass[10pt,journal,compsoc]{IEEEtran}
\ifCLASSOPTIONcompsoc
  \usepackage[nocompress]{cite}
\else
  \usepackage{cite}
\fi
\usepackage{epsfig}
\usepackage{graphicx}
\usepackage{amsmath}
\usepackage{amssymb}
\usepackage{bbding}
\usepackage{float} 
\usepackage{ulem}
\usepackage{booktabs}
\usepackage{colortbl}
\usepackage{xcolor}
\usepackage{subfigure}
\usepackage{lipsum}
\usepackage{algorithm}
\usepackage{algorithmic}
\usepackage{setspace}
\usepackage{ragged2e}
\usepackage{hyperref}

\definecolor{myblue}{RGB}{0,176,240}
\definecolor{myred}{RGB}{192,80,70}
\definecolor{purple}{RGB}{112,48,160}
\definecolor{myorange}{RGB}{255,192,0}
\ifCLASSINFOpdf
\else
\fi
\begin{document}
%
\title{360 Layout Estimation via Orthogonal Planes Disentanglement and Multi-view Geometric Consistency Perception}


%
%
%
%

\author{Zhijie~Shen,
        Chunyu~Lin,
        Junsong Zhang,
        Lang~Nie,
        Kang~Liao,
        Yao~Zhao,~\IEEEmembership{Fellow,~IEEE}
\IEEEcompsocitemizethanks{\IEEEcompsocthanksitem Corresponding author: Chunyu Lin
\IEEEcompsocthanksitem Zhijie Shen, Chunyu Lin, Junsong Zhang, Lang Nie, and Yao Zhao are with the Institute of Information Science, Beijing Jiaotong University, Beijing 100044, China, and also with the Beijing Key Laboratory of Advanced Information Science and Network Technology, Beijing 100044, China 
(e-mail: zhjshen@bjtu.edu.cn, cylin@bjtu.edu.cn, jszhang@bjtu.edu.cn, nielang@bjtu.edu.cn, yzhao@bjtu.edu.cn.)
\IEEEcompsocthanksitem Kang Liao is with the School of Computer Science and Engineering, Nanyang Technological University, Singapore (e-mail: kang.liao@ntu.edu.sg).}
}

\IEEEtitleabstractindextext{
\begin{abstract}
\justifying
Existing panoramic layout estimation solutions tend to recover room boundaries from a vertically compressed sequence, yielding imprecise results as the compression process often muddles the semantics between various planes. Besides, these data-driven approaches impose an urgent demand for massive data annotations, which are laborious and time-consuming. For the first problem, we propose an orthogonal plane disentanglement network (termed DOPNet) to distinguish ambiguous semantics. DOPNet consists of three modules that are integrated to deliver distortion-free, semantics-clean, and detail-sharp disentangled representations, which benefit the subsequent layout recovery. For the second problem, we present an unsupervised adaptation technique tailored for horizon-depth and ratio representations. Concretely, we introduce an optimization strategy for decision-level layout analysis and a 1D cost volume construction method for feature-level multi-view aggregation, both of which are designed to fully exploit the geometric consistency across multiple perspectives. The optimizer provides a reliable set of pseudo-labels for network training, while the 1D cost volume enriches each view with comprehensive scene information derived from other perspectives. Extensive experiments demonstrate that our solution outperforms other SoTA models on both monocular layout estimation and multi-view layout estimation tasks. Cobe can be available at~\url{https://github.com/zhijieshen-bjtu/MV-DOPNet}.
\end{abstract}

}

\maketitle

\IEEEdisplaynontitleabstractindextext

%
\IEEEpeerreviewmaketitle
\IEEEraisesectionheading{\section{Introduction}\label{sec:introduction}}
\IEEEPARstart{I}{ndoor} panoramic layout estimation refers to reconstructing 3D room layouts from omnidirectional images. Since the panoramic vision system captures the whole-room contextual information, we can estimate the complete room layout with a single panoramic image. However, the 360 field-of-view (FoV) inherent to panoramas introduces significant distortions that intensify with latitude, producing adverse effects on the perception of layout structures. Furthermore, current room-layout models are predicated on the Manhattan world assumption and trained in a supervised manner. For complex situations like non-cuboid rooms, these models might encounter undesired challenges, especially in the absence of annotations. In view of this, we revisit this field from two perspectives: developing an explainable fundamental model and constructing an unsupervised adaption strategy with multi-view geometric consistency.  

\begin{figure}[!t]
  \centering
  \includegraphics[width=0.5\textwidth]{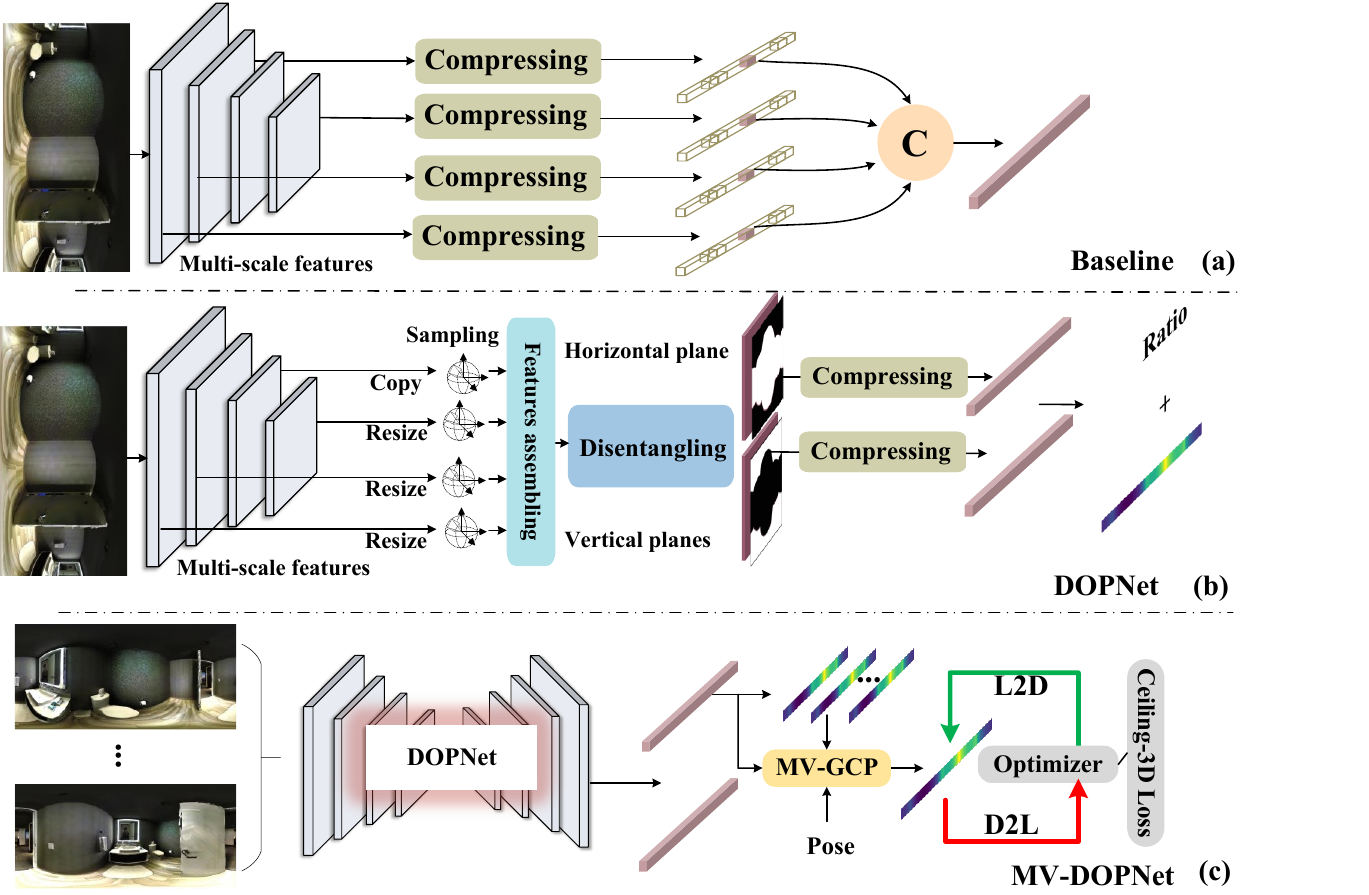} 
  \caption{(a) The commonly used architecture. (b) DOPNet. (c) MV-DOPNet.
  Compared with the traditional pipeline, ours has the following advantages: (1) Disentangling the orthogonal planes to get distortion-free, and semantic-clean, detail-sharp representations. (2) Leveraging multi-view geometric consistency at the feature level (MV-GCP module). (3) Optimization strategy suitable for horizon-depth/ratio layout representations.
  } 
  \label{fig:newarchi}
  \vspace{-0.2cm}
\end{figure}
Previous approaches \cite{sun2019horizonnet,sun2021hohonet,Jiang_2022_CVPR,pintore2020atlantanet,wang2021led} prefer to estimate the layouts from a 1D sequence. They advocate compressing the extracted 2D feature maps in the height dimension to obtain a 1D sequence, of which every element shares the same distortion magnitude (Fig. \ref{fig:newarchi}a). However, such a compressed representation does not eliminate panoramic distortions; instead, it merely distributes the distortion uniformly. Moreover, it roughly mixes the vertical and horizontal planes together, confusing the semantics of different planes that are crucial for layout estimation.  

To address the above problems, we propose a novel architecture (Fig.\ref{fig:newarchi}b) that disentangles the orthogonal planes to obtain distortion-free, semantics-clean, and detail-sharp horizon-depth/ratio representations. We termed the proposed architecture DOPNet, which comprises three modules to disentangle the inherent 1D representation step-by-step. Firstly, we propose a distortion-aware feature aggregation module to assemble cross-scale features without distortion. Secondly, we propose to pre-segment the vertical plane (\textit{i.e.}, walls) and horizontal planes (\textit{i.e.}, floors and ceilings) from the whole-room contextual information to incorporate 3D geometric cue to the layout model. Given the weak symmetry properties in general layout rooms, a soft-flipping fusion strategy with learnable offsets is designed to facilitate this process. 
The segmented orthogonal planes are then compressed into two 1D sequences. Considering the potential inaccuracy of pre-segmentation, in the third step, we present a sequence reconstruction module to carry out intra- and inter-sequence interaction. The horizon-depth and ratio are then separately regressed from the vertical and horizontal sequences to avoid mutual interference. 

On the other hand, we devise an unsupervised adaptation strategy for multi-view room layout without annotations, introducing both a horizon-depth optimizer and a multi-view perception module (Fig.\ref{fig:newarchi}{\color{red}c}, termed MV-DOPNet). The proposed optimizer, to our knowledge, stands as a pioneering step for horizon-depth representation in the field. It addresses the direct multi-view horizon-depth alignment challenge by shifting the optimization objective from depth to 2D layout boundaries. However, optimization horizon-depth alone brings a mismatch with the unoptimized ratio. To this end, we introduce a ceiling-3D loss function, which implicitly constrains the ratio via ceiling-depth (converted from horizon-depth), ensuring that the optimization process also benefits the ratio representation. Furthermore, we develop a multi-view geometric consistency perceptron module. It employs a coarse horizon-depth sequence combined with differentiable polar coordinate warping to map different features from other views into a unified reference view to construct a 1D cost volume. It explicitly integrates multi-view layout geometry into the model, as well as ensures a more comprehensive and accurate representation.

In this enhanced version of our previous work\cite{Shen_2023_CVPR}, we explore unsupervised multi-view layout estimation. We introduce a novel optimization approach and a 1D cost volume construction method for horizon-depth/ratio layout representation. Our focus is on leveraging multi-view geometric consistency at both decision and feature levels.
 
Finally, we conduct extensive experiments on four popular datasets --- MatterportLayout \cite{zou2021manhattan}, Zind \cite{cruz2021zillow}, Stanford 2D-3D \cite{armeni2017joint}, and PanoContext \cite{zhang2014panocontext} to demonstrate the effectiveness of DOPNet. Both qualitative and quantitative results show that the proposed solution outperforms other SoTA methods. Furthermore, experimental results on the MVL challenge dataset\footnote{\url{https://github.com/mvlchallenge/mvl_toolkit}} demonstrate our solution significantly outperforms the SoTA methods. Our contributions can be summarized as follows:
\begin{itemize} 
    \item We propose to disentangle ambiguous semantics to obtain spotless (distortion-free, semantics-clean, and detail-sharp) horizon-depth/ratio representations, which benefit the subsequent layout recovery. 
    \item We incorporate two mutual transformative functions to a horizon-depth optimization strategy, complemented by a ceiling-3D loss to share the optimization gains and enhance network fine-tuning.       
    \item We develop a multi-view perception to enhance latent features of the reference view with feature-level geometric consistency, digging into multi-view layout geometry represented by a 1D cost volume. 
    \item On popular benchmarks, our solution outperforms other SoTA schemes on both monocular layout estimation and multi-view layout estimation tasks.
\end{itemize}
 
\section{Related Work}
\subsection{Supervised 360 Room Layout Estimation}
\label{sec2_1}
Based on Manhattan World assumption \cite{coughlan1999manhattan} that all walls and floors are aligned with global coordinate system axes and are perpendicular to each other, many approaches utilize convolutional neural networks (CNNs) to extract useful features to improve layout estimation accuracy. Specifically, Zou $et$ $al.$ \cite{zou2018layoutnet} propose LayoutNet to predict the corner/boundary probability maps directly from the panoramas and then optimize the layout parameter to generate the final predicted results.
Furthermore, they annotate the Stanford 2D-3D dataset \cite{armeni2017joint} with awesome layouts for training and evaluation.
Yang $et$ $al.$ \cite{yang2019dula} propose Dula-Net to predict a 2D floor plane semantic mask from both the equirectangular view and the perspective view of the ceilings.
The modified version, LayoutNet v2, and Dula-Net v2, which have better performance than the original version, are presented by Zou $et$ $al.$ \cite{zou2021manhattan}.
Fernandez $et$ $al.$ \cite{fernandez2020corners} present to utilize equirectangular convolutions (EquiConvs) to generate corner/edge probability maps.
Sun $et$ $al.$ propose HorizonNet \cite{sun2019horizonnet} and HoHoNet \cite{sun2021hohonet} to simplify the layout estimation processes by representing the room layout with a 1D representation.
Besides, they use Bi-LSTM and multi-head self-attention to build long-range dependencies and refine the 1D sequences.
\begin{figure*}[t]
  \centering
  \includegraphics[width=\textwidth]{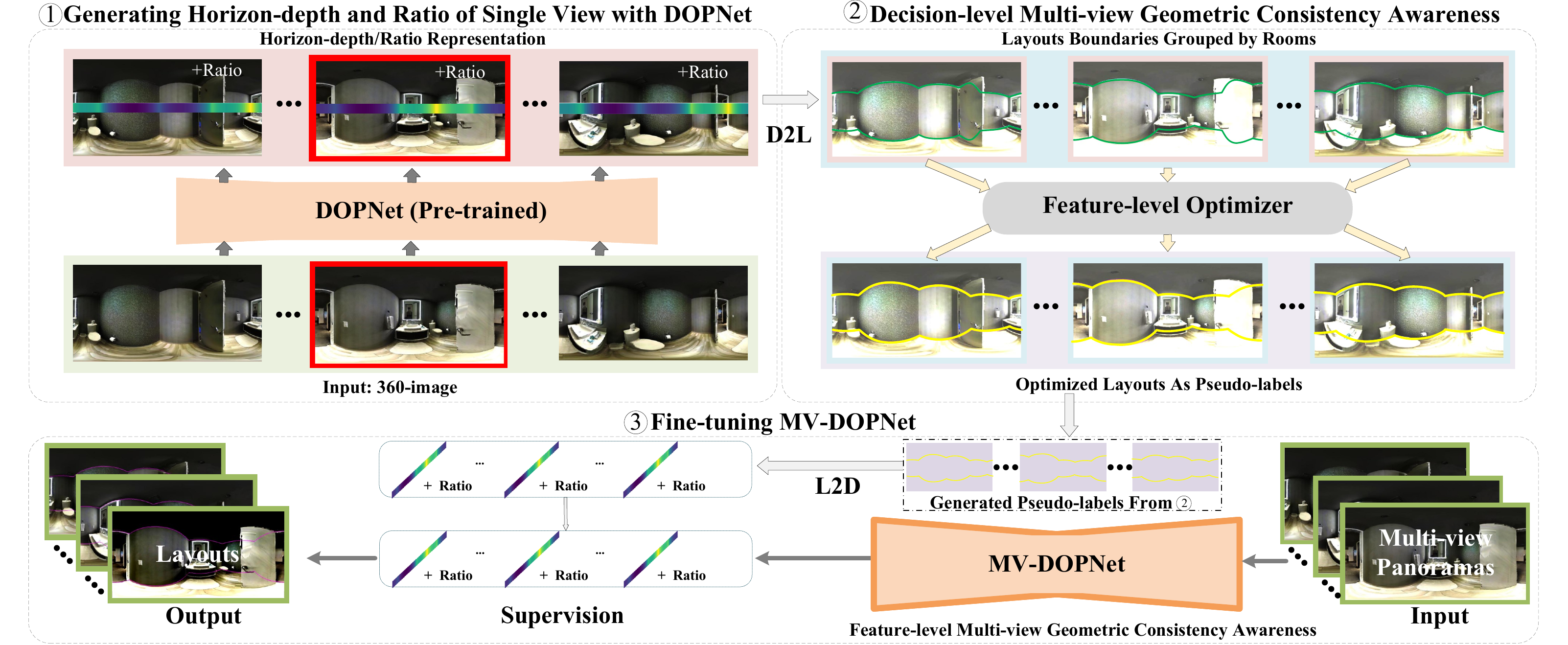} 
  \caption{Workflow of the proposed unsupervised multi-view layout framework. Our approach adopts panoramic images as input and predicts the corresponding horizon-depth map and the height (ratio). Then they are employed to recover the layout boundaries via D2L transformation. We leverage the multi-view layout consistency to optimize the boundaries and convert them back to the original format via L2D transformation. Finally, the optimized pseudo-labels are employed to fine-tune the MV-DOPNet (extending DOPNet by designing a multi-view geometric consistency perceptron module).} 
  \label{fig:doppluse}
\vspace{-0.2cm}
\end{figure*}
Recently, several approaches have adopted this 1D representation and achieved impressive performance. For example, Rao $et$ $al.$ \cite{rao2021omnilayout} establish their network based on HorizonNet \cite{sun2019horizonnet}. They replace standard convolution operation with spherical convolution to reduce distortions and take Bi-GRU to reduce computational complexity.
Wang $et$ $al.$ \cite{wang2021led} combine the geometric cues across the layout and propose LED2-Net that reformulates the room layout estimation as predicting the depth of walls in the horizontal direction.
Not constrained to Manhattan scenes, Pintore $et$ $al.$ \cite{pintore2020atlantanet} introduce AtlantaNet to predict the room layout by combining two projections of the floor and ceiling planes.
Driven by the self-attention mechanism, many transformer-based methods \cite{yuan2021tokens, wang2021pyramid, liu2021swin} proposed to build long-range dependencies. For example, Jiang $et$ $al.$ \cite{Jiang_2022_CVPR} represent the room layout by horizon depth and room height and introduce a Transformer to enforce the network to learn the geometric relationships. However, the popular 1D representations could confuse the semantics of different planes, thus causing the layout estimation challenging. We disentangle this 1D representation by pre-segmenting orthogonal planes. 

\subsection{Unsupervised 360 Room Layout Estimation}
\label{sec2_2}
In new target domains, such as more general rooms, existing models' performance declines due to different viewpoints, lighting conditions, or occlusions. Addressing these issues typically requires labeled data for fine-tuning. Additionally, the diversity of indoor rooms needs even more labeled data for room layout models to adapt to new domains. Recent studies devote themselves to developing unsupervised adaptation approaches for the new target domain. However, room layout representation is inherently sparse and topological, making it challenging to apply unsupervised constraints used in dense prediction tasks directly. This sparse representation can easily lead to local optima when applying image reprojection losses. To address these challenges, existing methods rely on pre-training techniques to obtain initial layout estimates for the subsequent self-training operation.  
The SSLayout360\cite{tran2021sslayout360} innovatively applies the traditional self-training technique, typically utilized for classification or object detection, to geometric prediction tasks like layout estimation. Specifically, this approach employs a pre-trained Mean Teacher \cite{tarvainen2017mean} model to generate pseudo-labels for unlabeled data. These pseudo-labels are subsequently incorporated into the training of existing layout estimation models. Since this method emphasizes monocular unsupervised layout estimation, it neglects the geometric constraints that exist between different views within a single room. In contrast, 360-MLC \cite{solarte2022mlc} advocates for leveraging geometric consistency among multi-views within the same room to generate more accurate pseudo-labels. Recent studies \cite{wang2021led} have highlighted the potential advantages of employing differentiable depth rendering techniques in layout estimation. Through the integration of depth prediction loss, it becomes feasible to incorporate geometrical insights into the model. However, the 360-MLC method designed for layout boundaries is unsuitable for these advanced approaches. Besides, the 360-MLC method still depends on a monocular layout model without any multi-view perception designs during the fine-tuning process. To address the challenges, we investigate a horizon-depth customized optimization methodology for these advanced models and explore an effective strategy for fine-tuning these advanced schemes on unlabeled data. 
\begin{figure*}[t]
  \centering
  \includegraphics[width=\textwidth]{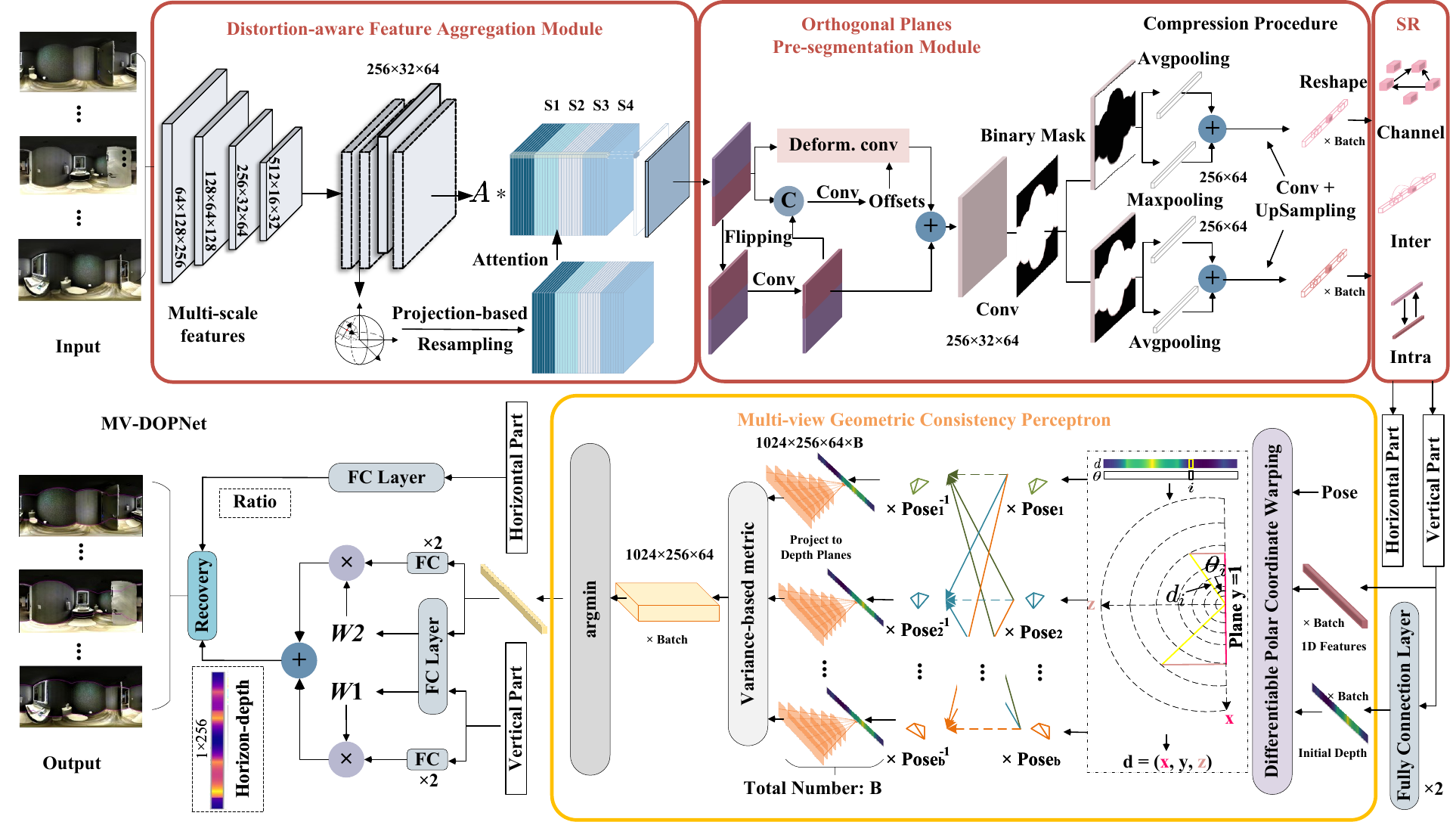} 
  \caption{MV-DOPNet. The feature assembling mechanism is designed to deal with distortions as well as integrate shallow and deep features. Then we pre-segment orthogonal planes to produce two 1D representations with distinguished plane semantics. A sequence reconstruction module is deployed to reconstruct the 1D representations. We design multi-view geometric consistency perceptron to incorporate multi-view insights. Our scheme leverages a sequence of panoramas with a resolution of 512×1024 as input and predicts the horizon-depth and the associated ratio value.} 
  \label{fig:framework}
\vspace{-0.2cm}
\end{figure*}
\subsection{360 Image Distortion Mitigation}
\label{sec2_3}
When converting spherical data to the equirectangular projection format, severe distortions are introduced. 
Some recent studies \cite{lee2020spherephd,xu2021spherical,zhuang2023spdet} focus on designing spherical customized convolutions to make the network adapt to panoramic distortions and offer superior results compared to standard convolutional networks.
Su $et$ $al.$ \cite{su2017learning} introduce SphConv to remove distortions by changing the regular planar convolutions kernel size towards each row of the 360 images.
Cohen $et$ $al.$ \cite{cohen2018spherical} propose a spherical CNN that transforms the domain space from Euclidean S2 space to a SO(3) group to reduce the distortion and encodes full rotation invariance in their network. 
Coors $et$ $al.$ \cite{coors2018spherenet} addresses it by defining the convolution kernels on the tangent plane and keeping the convolution domain consistent in each kernel. Therefore, the non-distorted features can be extracted directly on 360 images.
Rao $et$ $al.$ \cite{rao2021omnilayout} first applied spherical convolution operation to room layout task and showed significant improvements. Different from the above approaches, Shen $et$ $al.$ \cite{shen2022panoformer} propose the first panorama Transformer (PanoFormer) and divide the patches on the tangent plane to remove the negative effect of distortions. However, many of the latest studies in layout estimation tend to overlook distortions, operating under the assumption that 1D representations effectively address these issues. On the contrary, we propose a feature assembling mechanism with cross-scale distortion awareness to deal with distortions. Extensive experimental results demonstrate the necessity of handling distortions before generating 1D representations.
\section{Approach}
\label{section3}
\subsection{Overview}
As illustrated in Fig. \ref{fig:doppluse}, our proposed framework is tailored for unlabeled data from general room environments. This framework includes pseudo-label generation, multi-view perception, and model fine-tuning. We outline the sequential steps to the framework's design as follows.

\noindent \textbf{Generating pseudo-labels.} For the unlabeled MVL data, we first follow the previous method \cite{tran2021sslayout360}\cite{solarte2022mlc} to generate pseudo-labels. Specifically, we employ our DOPNet to generate initial horizon-depth maps and ratios for the multiple views. 

\noindent \textbf{Multi-view Geometric Consistency Awareness.} These horizon-depth sequences are grouped by rooms and undergo refinement using Algorithm \ref{alg:algorithm}  to create pseudo-labels for fine-tuning. Additionally, during the fine-tuning period, we present MV-DOPNet, enabling feature-level multi-view insights of layout models.

\noindent \textbf{Fine-tuning with Pseudo-labels.} The optimized pseudo-labels are employed to fine-tune the proposed multi-view layout model (MV-DOPNet). Our model, upon receiving multi-view panoramas as input, generates the horizon-depth and ratio for each view, which are then utilized to recover the corresponding room layouts. With our customized constraints, this fine-tuning process taps into a broader data spectrum, enhancing MV-DOPNet's inference and generalization capabilities. 

\subsection{Disentangle Orthogonal Planes}
We first briefly review the details of DOPNet \cite{Shen_2023_CVPR}. As illustrated in Fig. \ref{fig:framework} (circled in red), the proposed framework consists of three stages. Specifically, we adopt a backbone (basically ResNet) to extract a series of features at 4 different scales from a single panorama. Then a distortion-aware feature aggregation module is designed to fuse the multi-scale features and free them from panoramic distortions. In the next stage, we introduce a soft-flipping fusion strategy to explore the weak symmetry properties between the floor boundary and the ceiling boundary. Subsequently, we segment the vertical and horizontal planes from the fused features followed by vertical compression, yielding two 1D sequences. In the last stage, the segmented features are reconstructed to be more discriminative and informative. Finally, the horizon-depth map and ratio value are regressed from the precise 1D representations.

\noindent \textbf{Distortion-aware Feature Aggregation Module.}
In previous works \cite{sun2019horizonnet,sun2021hohonet,wang2021led,Jiang_2022_CVPR}, researchers prefer to follow the architecture of HorizonNet \cite{sun2019horizonnet} to extract multi-scale feature maps of an equirectangular image. These feature maps are merged together via vertical compression and concatenation \cite{lin2017feature}. However, the popular architecture just concatenates multi-scale features and neglects the inherent distortion distribution, resulting in an inferior panoramic feature extraction capability. 
To address the above issues, we propose a novel distortion-aware feature aggregation module to deal with distortions, as well as ingeniously integrate shallow and deep features.

Motivated by \cite{fernandez2020corners,shen2022panoformer}, we first polymerize the most relevant features via a virtual tangent plane on the panoramic sphere projection. Based on the projection formula \cite{shen2022panoformer,shen2021distortion,coors2018spherenet}, we can get the distortion-free sampling positions in the 2D feature maps. Then we employ the learnable offsets to adjust the sampling locations adaptively and obtain gathered features from the distorted features. Then we select the $3^{rd}$ scale as the reference scale and resize other feature maps to this scale. Specifically, we utilize the learnable sampling positions to gather the distortion-free feature maps for every feature with resized scales.
\begin{figure}[H]
  \centering
  \includegraphics[width=0.45\textwidth]{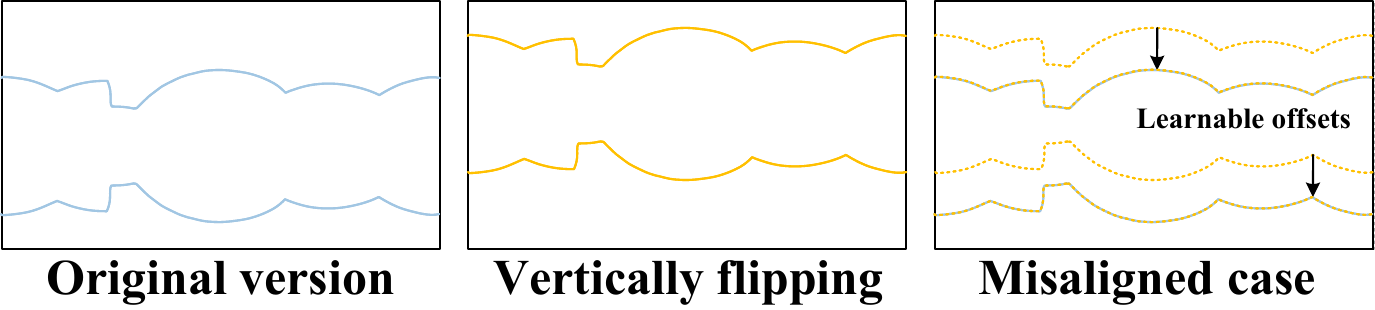} 
  \caption{Illustration of the misaligned case when leveraging the symmetry property.} 
  \label{fig:soft}
\vspace{-0.2cm}
\end{figure}
\noindent \textbf{Orthogonal Planes Pre-segmentation Module.} To explicitly capture the geometric cues, we propose to segment the vertical and horizontal planes from the whole scenario before compressing the 2D feature maps. Specifically, we pre-segment the horizon/vertical planes in advance to capture an explicit geometric cue in 3D space and encourage the network to learn a binary mask via a simple segmentation head. Following previous works \cite{sun2021hohonet,sun2019horizonnet}, we further vertically compress 2D feature maps to two 1D sequences with different plane semantics.

Besides, we observe that it is challenging to learn accurate layout boundaries in a complex scene due to the disturbance of furniture and illumination, \textit{e.g.}, occlusion. In Dula-Net\cite{yang2019dula}, Yang \textit{et al.} propose a dual-branch network, leveraging a perspective ceiling view branch to assist in accurate layout prediction. However, the ceiling-view solves wall occlusion issues, but it compounds the occlusion effect from the indoor decorations (\textit{e.g.}, furniture and lighting). Besides, Dula-Net aims to predict a plan probability map, while our goal is to estimate a horizon-depth map and an associated ratio value. The view conversion operation can result in inconsistent depth, hindering accurate horizon-depth estimation. To address these challenges, we propose to leverage the symmetry property of indoor rooms (\textit{e.g.}, the floor boundary and ceiling boundary are mostly symmetric in 3D space) to provide complementary information. We introduce a soft-flipping fusion strategy to exploit this symmetry property. We first vertically flip the feature map to get their symmetrical version. In fact, the floor boundary and ceiling boundary are not strictly symmetric in a general room scene  (illustrated in Fig. \ref{fig:soft}). To this end, we adopt a deformable convolution with 3×3 kernel size to adaptively adjust the symmetrical version. Next, we fuse the original feature and its ``soft" symmetrical version to provide more informative boundary cues.  

\noindent \textbf{Sequences Reconstruction Module.}
Previous segmentation and compression operations disrupt inherent dependencies in 2D features. To this end, we advocate generating discriminative channels, establishing solid long-range dependencies, and enhancing helpful residuals. 
Li $et$ $al.$ ~\cite{graph_reasoning} find that there are generally unconfined dependencies among channels, resulting in confused semantic cues. To this end, we employ graph convolution to enforce each channel to concentrate on distinguishing features. Different from normal pixel-wise~\cite{pixel_level} or object-wise~\cite{object_level} graphs, the introduced channel-wise one tends to guide the node to subtract the information from the neighbor nodes rather than aggregation. 
 
Moreover, we employ the standard self-attention to reconstruct the intra-sequence long-range dependencies. It captures global interactions to adapt to the large FoV of panoramas, contributing to better performance. To compensate for inevitable errors in pre-segmentation, we introduce cross-attention to provide the missing useful residuals. Then we add the learned residuals with the two 1D sequences to implement the reconstruction. After that, two separate fully connected layers are employed to output the predicted horizon depth $d \in \mathbb{R}^{B\times W}$ from the vertical plane sequence and the ratio value from the horizontal plane one, respectively.

\subsection{Multi-view Geometric Consistency Awareness}
\label{section_ext}
The strategies proposed in this section are specifically tailored for multi-view scenarios. We align the multi-view horizon-depths to fully harness the potential of geometric consistency across different views, thereby enhancing the precision of layout estimation tasks. Specifically, the alignment is performed in two steps: decision-level optimization and feature-level perception. At the decision level, estimated horizon depths are projected onto a 2D canvas for optimization, which avoids the issues of scale and shifting caused by changes in poses, lighting conditions, and occlusions across multiple views. In the feature-level perception, multi-view horizon-depths are initially aligned with the camera pose information. Subsequently, through fine-tuning, the model can leverage the complementary features from other views to deal with the occlusions and produce more accurate horizon-depths for each view. 
\subsubsection{Decision-level Optimization}
\textbf{Horizon-depth Optimization.} Our approach utilizes the advanced horizon-depth and ratio representation. However, under the assumption in \cite{wang2021led}, horizon-depth sequences generated from the single-view layout model (DOPNet) are not robust due to the varying lighting conditions and occlusions across different views. Directly optimizing depth using camera pose information is challenging. To this end, we propose to project these depth values onto a canvas, thereby shifting the optimization object from 3D space into 2D space. Specifically, we introduce the Depth-to-Layout (D2L) conversion, the reverse of the Layout-to-Depth (L2D) conversion \cite{wang2021led}, ensuring the method's compatibility within the advanced layout estimation framework. Following refinement, we employ the L2D transformation to maintain the consistency of the model's original representation schema. This entire cycle is described in Algorithm \ref{alg:algorithm}. Notably, we incorporate the calculation of standard deviations among multi-view horizon-depth maps to formulate the loss function.
\begin{algorithm}[tb]
\caption{Pseudo-labels Generation}
\label{alg:algorithm}
\textbf{Input}: Registered horizon-depth sequences $D$ within a room and the corresponding camera pose $P$ relative to the world coordinates\\
\textbf{Output}: Optimized pseudo-labels $\hat{D}$, standard deviation $\sigma$ 
\begin{algorithmic}[1] 
\FOR{ $d$ in $D$}
\STATE    horizon-depth to layout: $l$ = D2L($d$)
\STATE    layout in unified world coordinates: $l^{wc}$ = $l$×$p_d$
\ENDFOR
\STATE All the unified layouts: $L^{wc}$
\FOR{$p$ in $P$}
\FOR{$l^{wc}$ in $L^{wc}$}
\STATE   layout at the reference view (associated with $p$):\\ $l_p$ = $l^{wc}$×$(p)^{-1}$
\ENDFOR
\STATE All the layouts at the reference view: $L_p$.
\STATE $\hat{l_p} \longleftarrow$ select high-confidence layouts point by point within $L_p$ (majority principle)
\STATE $\sigma_p \longleftarrow$ standard deviation of $L_p$
\STATE layout to horizon-depth: $\hat{d_p}$ = L2D($\hat{l_p}$)
\ENDFOR
\STATE \textbf{return} $\hat{D}$, $\sigma$
\end{algorithmic}
\end{algorithm}

\noindent \textbf{Ceiling-3D Constraint for Ratio.} The layout model's output implies a correlation wherein the horizon-depth $d$ is related to the ceiling-depth $d^c$. Upon transforming the depth readings into 3D cartesian coordinates, we ascertain the spatial 3D points designated as $\mathbf{d} = (\mathbf{x}, \mathbf{y}, \mathbf{z})$ and $\mathbf{d}^c = (\mathbf{x}^c, \mathbf{y}^c, \mathbf{z}^c)$. These coordinates sustain a definable relationship, expressed by the following expression:
\begin{eqnarray}
\begin{array}{cc}
\mathbf{y}^c = -R \mathbf{y},
\end{array}
\end{eqnarray}
where $\mathbf{d}, \mathbf{d}^c \in \mathbb{R}^{B\times3\times W}$; $R$ indicates the predicted ratio value. Consequently, we formulate constraints on the ceiling depth rather than imposing direct constraints on the ratio. This improvement facilitates a more direct control over the model's geometric fidelity, thereby refining the overall accuracy of the layout estimation.

\subsubsection{Feature-level Perception} We introduce a multi-view geometric consistency perceptron, depicted in Fig. \ref{fig:framework} (highlighted in Orange), designed to integrate multi-view insights at the feature level by constructing 1D cost volumes.

\noindent\textbf{Differentiable Polar Coordinate Warping.} Given the width of the sequence $W$, we first calculate the longitude values $\theta$:
\begin{eqnarray}
\begin{array}{cc}   
\forall i \in \{0, \dots, W-1\} : \quad \theta = \left(\frac{i + 0.5}{W} - 0.5\right) 2\pi.
\end{array}
\end{eqnarray}
With the coarse horizon-depth $d$ generated, we apply a differentiable polar coordinate warping method to obtain its 3D coordinates $\mathbf{d} = (\mathbf{x}, \mathbf{y}, \mathbf{z})$ in the cartesian coordinate system:
\begin{equation} 
\left\{
	\begin{array}{ll}
		\mathbf{x}& =  d \sin \theta  \\
		\mathbf{y} & =  1\\
        \mathbf{z} & = d \cos \theta
	\end{array},
\right.
\end{equation}
where all the operations are element-wise. Afterward,  we align the 3D points $\mathbf{d}$ of other views to the reference view by employing the relative camera poses: 
\begin{eqnarray}
\begin{array}{cc}
\mathbf{d}^{align} = p^{ref} \times \mathbf{d},
\end{array}
\end{eqnarray}
where $p^{ref}$ indicates the matrix of the relative camera pose between the current view and the target view. 

\noindent\textbf{Depth Plane Sampling.} Given 64 depth planes which are defined linearly across the normalized depth range [0, 1], yielding intervals 1/64, masks $\mathbf{m}$ are created for each depth plane by checking if the z-coordinate of the aligned points $\mathbf{d}^{align}$ lies within the depth plane. For each view, we apply the mask to the expanded features $\mathbf{f} \in \mathbb{R}^{B\times C \times W \times 64}$ to get the volumes:
\begin{eqnarray}
\begin{array}{cc}
\mathbf{v} = \mathbf{f} \odot \mathbf{m},
\end{array}
\end{eqnarray}
where $\odot$ represents the element-wise product. Eventually, for each reference view, we can get all the volumes $\mathbf{V} \in \mathbb{R}^{B\times C \times W \times 64 \times B}$.

\noindent\textbf{1D Cost Volumes.} Similar to \cite{yao2018mvsnet}, we employ the variance-based metric to aggregate $\mathbf{V}$ to one 1D cost volume $\mathbf{c}$. The process can be written as follows:
\begin{eqnarray}
\begin{array}{cc}
\mathbf{c} = (\sum\limits_{i = 1}^{B}\mathbf{V}_i - \bar{\mathbf{V}}_i^2)/B,
\end{array}
\end{eqnarray}
where $\mathbf{c} \in \mathbb{R}^{B\times C \times W \times 64}$, and $\bar{\mathbf{V}}_i$ is the average volume among all features volumes. We finally apply the \textit{argmin} operation among the depth planes to get a precise 1D feature: 
\begin{eqnarray}
\begin{array}{cc}
\mathbf{f^c} = argmin_\mathbf{c} c,
\end{array}
\end{eqnarray}
where $\mathbf{f^c} \in \mathbb{R}^{B\times C \times W}$. Then we employ a fully connected layer to get a depth sequence $d^{cost}$ from $\mathbf{f^c}$.
Finally, we fuse the two depths to ensure an optimized depth representation. The final depth can be expressed as:
\begin{eqnarray}
\begin{array}{cc}
d^{final} = W_1 d^{cost} + W_2 d, 
\end{array}
\end{eqnarray}
where $d^{cost}$ represents the depth obtained from the cost volume; \( W_1 \) and \( W_2 \) are the weights determined by the fully connected layers.
 
\subsection{Loss Function}
\subsubsection{Objective Function for Pre-training}
Our objective function consists of two parts: one for room layout estimation and the other for plane segmentation. For the first part, we strictly follow \cite{Jiang_2022_CVPR} and denote it as $\mathcal{L}_{lay}$. For the segmentation part, we apply binary cross-entropy loss $\mathcal{L}_{seg}$ as follows:
\begin{eqnarray}
\begin{array}{cc}   
\mathcal{L}_{seg}= -\frac{1}{N} \sum\limits _{i=1}^{N}(d_{i}\log_{}{\hat{d_{i}}}+(1-d_{i})\log_{}{(1-\hat{d_{i}}))},
\end{array}
\end{eqnarray}
where $d_i$ is the ground truth of the generated segmentation label, and $\hat{d}_i$ is the predicted value. Ultimately, we formulate the objective function as follows:
\begin{eqnarray}
\begin{array}{cc}   
\mathcal{L}= \mu \mathcal{L}_{seg} + \mathcal{L}_{lay},
\end{array}
\end{eqnarray}
where $\mu$ is set to 0.75 to balance different constraints.
\subsubsection{Objective Function for Fine-tuning}
The fine-tuning phase of our model employs a loss function comprised of four distinct elements: normals, gradients of normals, horizon-depth, and ratio. Firstly, we employ \( \sigma \) which is calculated in Algorithm \ref{alg:algorithm} as weights to design the losses:
\begin{align}
d & = \frac{d}{\sigma^2},\\
\hat{d} & = \frac{\hat{d}}{\sigma^2},
\end{align}
where the operations are element-wise.  Then we carefully follow the methodology outlined in \cite{Jiang_2022_CVPR} for the first three components, symbolized by \( \mathcal{L}_{n}^w \), \( \mathcal{L}_{g}^w \), and \( \mathcal{L}_{d}^w \). The subsequent ratio loss (ceiling-3D loss) is as follows,
\begin{align}
\mathcal{L}_{r} = \frac{1}{3N} \sum\limits_{i=1}^{3} \sum_{j=1}^{N} \left| \frac{\mathbf{d}^c_{i,j}-\hat{\mathbf{d}^c}_{i,j}}{\sigma_{i,j}^2} \right|,
\end{align}
where $\mathbf{d}^c_{i,j}$ indicates the ceiling-depth converted from the pseudo-label, and $\hat{\mathbf{d}^c}_{i,j}$ is the one converted from the predicted horizon-depth.
Combining all these components, the cumulative objective function is given by:
\begin{align}
\mathcal{L} = \lambda_1 (\mathcal{L}_{n}^w + \mathcal{L}_{g}^w) + \lambda_2\mathcal{L}_{d}^w + \lambda_3 \mathcal{L}_{r}.
\end{align}
In our implementation, we empirically set \( \lambda_1 \) to 0.1, \( \lambda_2 \) to 0.9, and \(\lambda_3\) to 0.08 to ensure a harmonious balance between various constraints.
 
\section{Experiments}
\label{section4}
In this section, we validate the effectiveness of our solution and compare it with existing SoTA approaches on five popular datasets.
Concretely, we conduct experiments on a single GTX 3090 GPU, and the batch size is set to 16 for training. The proposed approach is implemented with PyTorch. We choose Adam~\cite{kingma2015adam} as the optimizer and keep the default settings. The initialized learning rate is $1\times10^{-4}$. As in previous works \cite{sun2019horizonnet,Jiang_2022_CVPR}, we adopt standard left-right flipping, panoramic horizontal rotation, luminance change, and pano stretch for data augmentation during training. Furthermore, we compare our framework against the  SoTA baseline method, 360-MLC. To ensure an equitable assessment, we retain the same parameters and settings.

\subsection{Datasets}
\noindent\textbf{Monocular Layout Estimation Datasets.} Four datasets are used for our experimental validation: Stanford 2D-3D~\cite{armeni2017joint}, PanoContext~\cite{zhang2014panocontext}, MatterportLayout~\cite{zou2021manhattan} and ZInd~\cite{cruz2021zillow}. 

PanoContext~\cite{zhang2014panocontext} and Stanford 2D-3D~\cite{armeni2017joint} are two commonly used datasets for indoor panoramic room layout estimation that contain 514 and 552 cuboid room layouts, respectively. Especially,  Stanford 2D-3D~\cite{armeni2017joint} is labeled by Zou $et$ $al.$ \cite{zou2018layoutnet} and has a smaller vertical FoV than other datasets. Besides, MatterportLayout \cite{zou2021manhattan} is also annotated by Zou $et$ $al.$ \cite{zou2021manhattan}, which contains 2,295 general room layouts. The final ZInd \cite{cruz2021zillow} dataset includes cuboid, general Manhattan, non-Manhattan, and non-flat ceilings layouts, which better mimic the real-world data distribution. The split of ZInd \cite{cruz2021zillow} consists of 24,882 (for training), 3,080 (for validation), and 3,170 (for testing) panoramas. We use the official splits of the four datasets for the monocular layout estimation task and follow the same training/validation protocols as previous works\cite{Jiang_2022_CVPR} for a fair comparison.

\noindent\textbf{Multi-view Layout Estimation Datasets.}
To validate the efficiency of our proposed framework, MV-DOPNet, we introduce a dataset specifically tailored for the Multi-View Layout Challenge. This challenge dataset is divided into two subsets: MP3D-FPE and HM3D-MVL. MP3D-FPE originates from the Matterport3D \cite{chang2017matterport3d} dataset and is amassed using the MINOS \cite{savva2017minos} simulator. It contains 50 diverse scenes, aggregating 687 rooms in all. Conversely, HM3D-MVL is a recently curated dataset derived from HM3D \cite{ramakrishnan2021habitat}, utilizing the Habitat \cite{savva2019habitat} simulator. Echoing the methodology in MP3D-FPE, for HM3D-MVL, the organizer replicated user scanning patterns to encapsulate more genuine camera movement dynamics. For a fair comparison, we employ the training splits from the challenge phase for model training and leverage the pilot splits to evaluate the performance. Notably, since the test splits of the MVL dataset from the challenge phase are not publicly released, we do not use them. 
\subsection{Evaluation Metrics}
We employ the commonly used standard evaluation metrics for a fair comparison, including corner error (CE), pixel error (PE), intersection over the union of floor shapes (2DIoU), and 3D room layouts (3DIoU). Among them, 3DIoU yields a better reflection of the accuracy of the layout estimation in 3D space. RMSE and $\delta_{1.25}$ indicate the performance of depth estimation, \textit{e.g.} the horizon-depth map. 
\begin{table}[t]
 \caption{Quantitative comparison results with the current SoTA solutions evaluated
on Stanford 2D-3D and PanoContext\cite{cruz2021zillow} dataset.}
 \label{tab:cp1}
 \vspace{-0.2cm}
\begin{center}
 \begin{tabular}{l c c c}
  \toprule
  Method& 3DIoU(\%)& CE(\%)& PE(\%) \\
    \hline
   \multicolumn{4}{c}{Train on PanoContext + Whole Stnfd.2D3D datasets}\\
   \hline LayoutNetv2\cite{zou2021manhattan}&85.02&0.63& 1.79\\
   Dula-Netv2\cite{zou2021manhattan}&83.77&0.81& 2.43\\
    HorizonNet\cite{sun2019horizonnet}&82.63&0.74& 2.17\\  
    LGT-Net\cite{Jiang_2022_CVPR}&85.16&-& -\\
    LGT-Net [w/ Post-proc]\cite{Jiang_2022_CVPR}&84.94&0.69&2.07\\
    Ours&\textbf{85.46}&-&-\\
    Ours [w/ Post-proc]&85.00&\textbf{0.69}&2.13\\
     \hline
   \multicolumn{4}{c}{Train on Stnfd.2D3D + Whole PanoContext datasets}\\
   \hline 
   LayoutNetv2\cite{zou2021manhattan}&82.66&0.83& 2.59\\
   Dula-Netv2\cite{zou2021manhattan}&\textbf{86.60}&0.67& 2.48\\
    HorizonNet\cite{sun2019horizonnet}&82.72&0.69& 2.27\\ 
    AtlantaNet\cite{pintore2020atlantanet}&83.94&0.71&2.18\\
    LGT-Net\cite{Jiang_2022_CVPR}&85.76&-&-\\
    LGT-Net [w/ Post-proc]\cite{Jiang_2022_CVPR}&86.03&0.63&2.11\\
    Ours&85.47&-&-\\
    Ours [w/ Post-proc]&85.58&0.66&\textbf{2.10}\\
  \bottomrule
 \end{tabular}
\end{center}
\end{table}
\subsection{Cuboid Room Results} 
We follow LGT-Net \cite{Jiang_2022_CVPR} to use the combined dataset scheme mentioned in Zou $et$ $al.$ \cite{zou2021manhattan} to evaluate our network on cuboid datasets. We denote the combined dataset that contains training splits of PanoContext \cite{cruz2021zillow} and whole Stanford 2D-3D datasets as ``Train on PanoContext + Whole Stnfd.2D3D datasets" in Tab. \ref{tab:cp1}. Similarly, ``Train on Stnfd.2D3D + Whole PanoContext" in Tab. \ref{tab:cp1} represents that we train our network on the combined dataset that contains training splits of Stanford 2D-3D and the whole PanoContext dataset. The scheme is commonly used in previous works \cite{zou2018layoutnet,Jiang_2022_CVPR}. We also report the results with a post-processing strategy in DuLa-Net \cite{yang2019dula} that is denoted as "Ours [w/ Post-proc]".

\noindent\textbf{Comparison Results.} We exhibit the quantitative comparison results on cuboid room layouts in Tab. \ref{tab:cp1}. From the first group in Tab. \ref{tab:cp1}, we can observe that our approach outperforms all the other SoTA schemes with respect to 3DIoU. But from the second group, Dula-Net v2 \cite{zou2021manhattan} offers better 3DIoU than ours. Dula-Net v2 \cite{zou2021manhattan} employs a perspective view (\textit{i.e.}, cubemap) that is more effective for panoramas with small vertical FoV. However, once applied to more general room layout datasets (MatterportLayout\cite{zou2021manhattan} dataset), its performance degrades significantly. On the contrary, the proposed approach shows more general performance. 
\begin{table}[t]
 \caption{Quantitative comparison results with the current SoTA solutions evaluated
on MatterportLayout\cite{zou2021manhattan} dataset and ZInD\cite{cruz2021zillow} dataset.}
 \label{tab:cp2}
\begin{center}
 \begin{tabular}{p{2.5cm} p{1.2cm} p{1.2cm} p{0.7cm} p{0.7cm}}
  \toprule
  Method& 2DIoU(\%) &  3DIoU(\%)& RMSE& $\delta_{1}$ \\
    \hline
    \multicolumn{4}{c}{MatterportLayout datasets}\\
    \hline
  LayoutNetv2\cite{zou2021manhattan}&78.73&75.82& 0.258&0.871\\
   Dula-Netv2\cite{zou2021manhattan}&78.82&75.05& 0.291&0.818\\
    HorizonNet\cite{sun2019horizonnet}&81.71&79.11& 0.197&0.929\\    
    AtlantaNet\cite{pintore2020atlantanet}&82.09&80.02& -&-\\
    HoHoNet\cite{sun2021hohonet}&82.32&79.88& -&-\\
    LED$^{2}$-Net\cite{wang2021led}&82.61&80.14& 0.207&0.947\\
    DMH-Net \cite{zhao20223d}&81.25&78.97& -&0.925\\
    LGT-Net\cite{Jiang_2022_CVPR}&83.52&81.11& 0.204&\textbf{0.951}\\
    Ours&\textbf{84.11}&\textbf{81.70}& \textbf{0.197}&0.950\\
    \hline
    \multicolumn{4}{c}{ZInD datasets}\\
    \hline
    HorizonNet\cite{sun2019horizonnet}&90.44&88.59& 0.123&0.957\\    
    LED$^{2}$-Net\cite{wang2021led}&90.36&88.49& 0.124&0.955\\
    LGT-Net\cite{Jiang_2022_CVPR}&91.77&89.95& 0.111&0.960\\
    Ours&\textbf{91.94}&\textbf{90.13}&\textbf{0.109}&0.960\\
  \bottomrule
 \end{tabular}
\end{center}
\end{table}
\begin{figure*}[htbp]
	\centering
	\subfigure[Our qualitative results on Stanford 2D-3D \cite{armeni2017joint} dataset.]{
		\begin{minipage}[t]{\linewidth}
			\centering
			\includegraphics[width=0.97\textwidth]{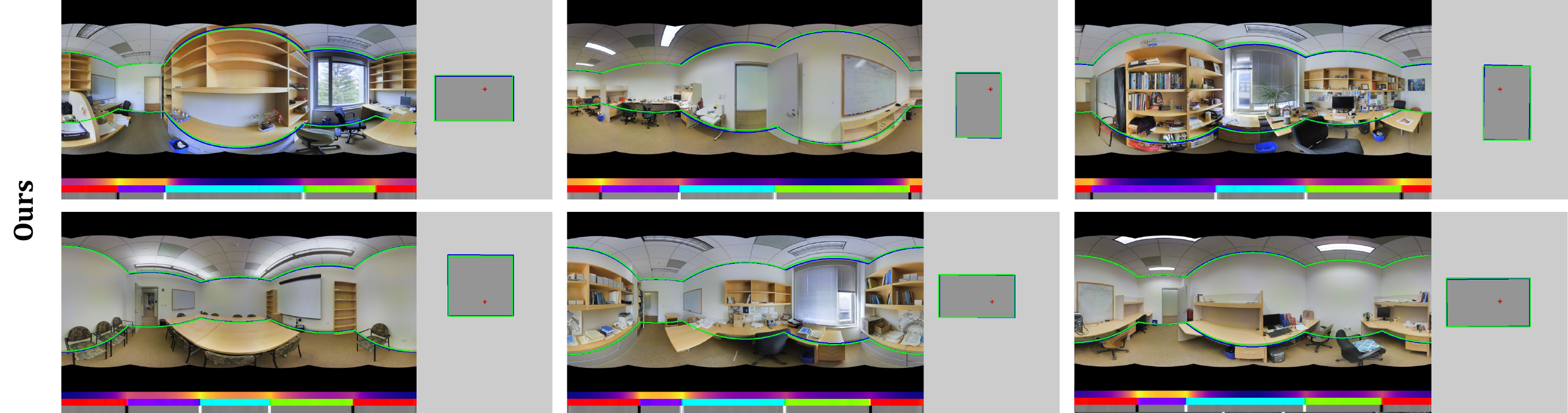}
		\end{minipage}
	}
    \subfigure[Our qualitative comparison on PanoContext \cite{zhang2014panocontext} dataset.]{
		\begin{minipage}[t]{\linewidth}
			\centering
			\includegraphics[width=0.97\textwidth]{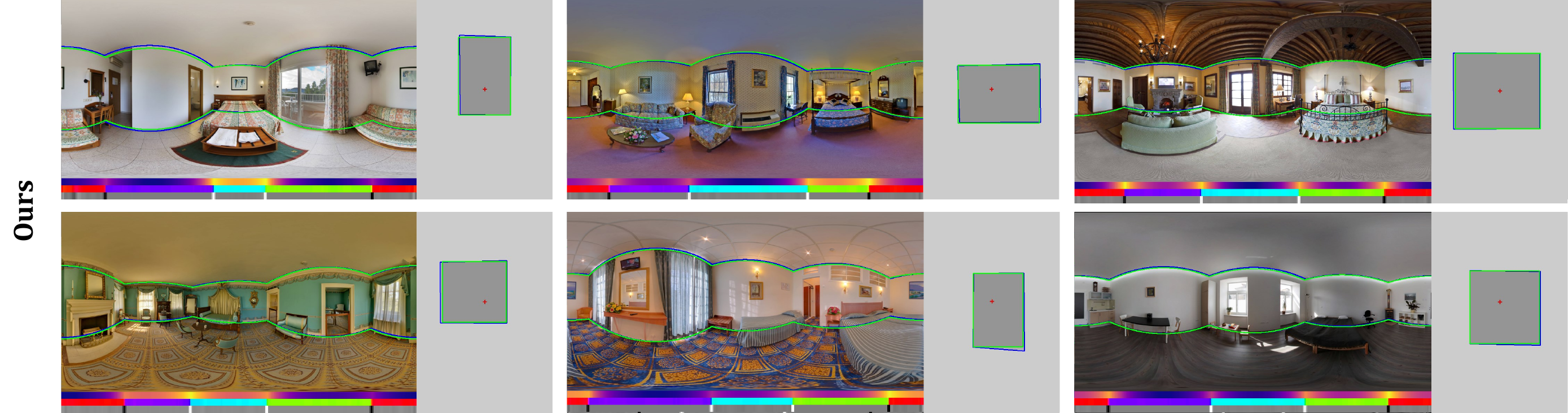}
		\end{minipage}
	}
	\caption{Qualitative comparison results evaluated on cuboid layout datasets, Stanford 2D-3D \cite{armeni2017joint} and PanoContext \cite{zhang2014panocontext}. The boundaries of the room layout on a panorama are shown on the left and the floor plan is on the right. Ground truth is best viewed in blue lines and the prediction in green. The predicted horizon depth, normal, and gradient are visualized below each panorama, and the ground truth is in the first row.}
	\label{fig:qulicuboid}
\vspace{-0.2cm}
\end{figure*}
\begin{figure*}[htbp]
	\centering
	\subfigure[Qualitative comparison on MatterportLayout \cite{zou2021manhattan} dataset.]{
		\begin{minipage}[t]{\linewidth}
			\centering
			\includegraphics[width=0.97\textwidth]{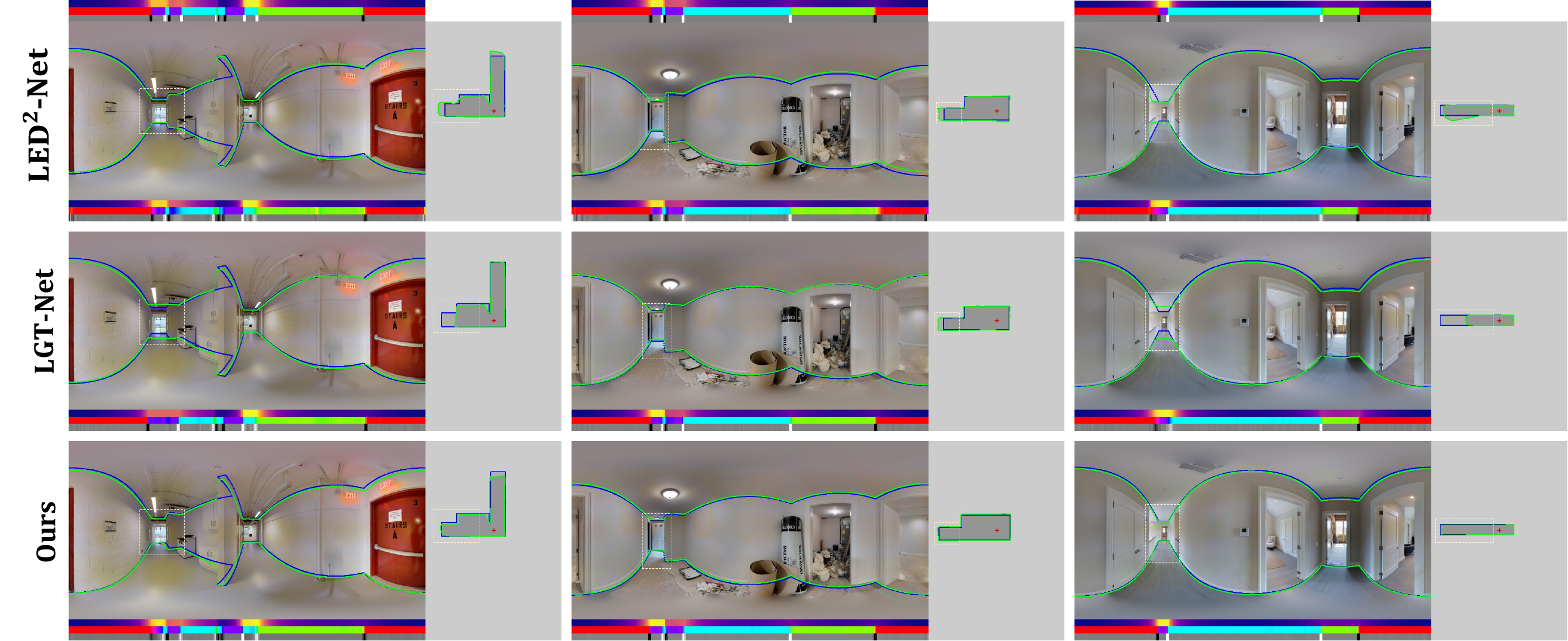}
		\end{minipage}
	}
    \subfigure[Qualitative comparison on ZInD \cite{cruz2021zillow} dataset.]{
		\begin{minipage}[t]{\linewidth}
			\centering
			\includegraphics[width=0.97\textwidth]{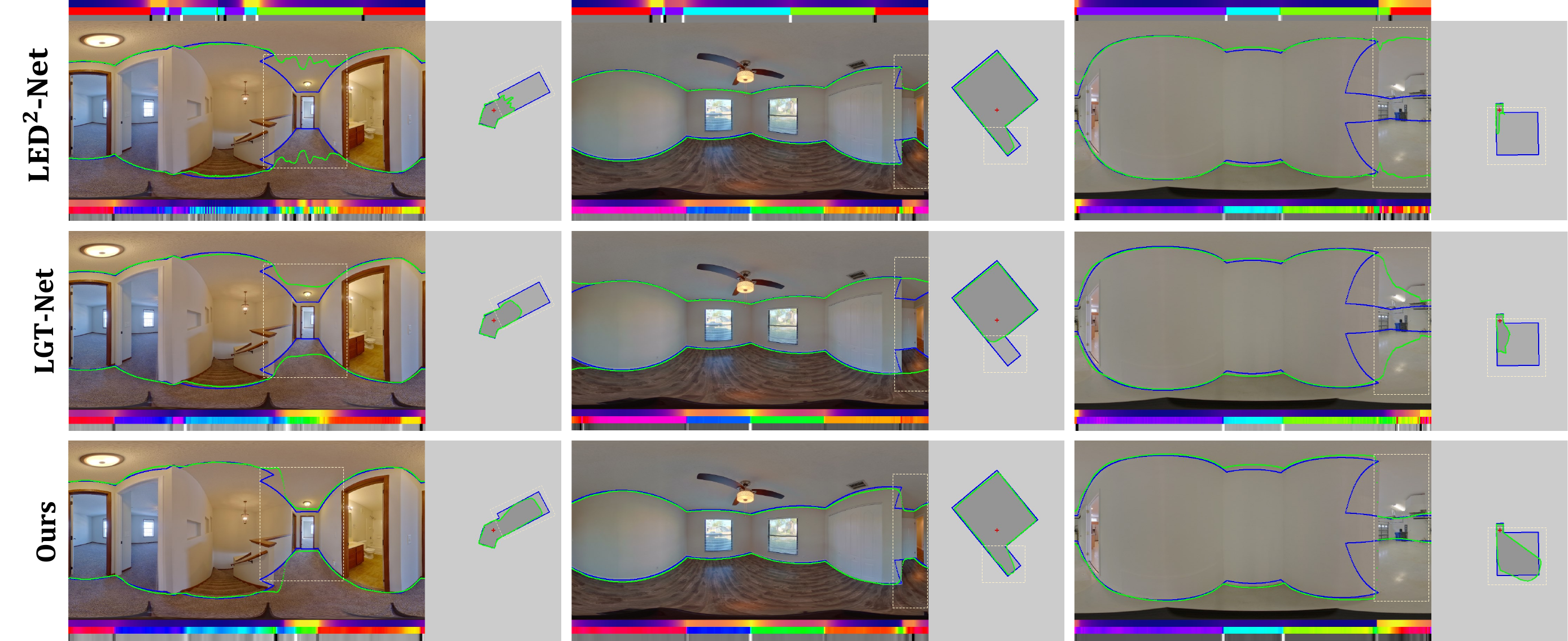}
		\end{minipage}
	}
	\caption{Qualitative comparison results evaluated on general layout datasets, MatterportLayout \cite{zou2021manhattan} and ZInD \cite{cruz2021zillow}. We compare our method with LED$^2$-Net \cite{wang2021led} and LGT-Net \cite{Jiang_2022_CVPR}. All the compared methods did not employ the post-processing strategy. The boundaries of the room layout on a panorama are shown on the left and the floor plan is on the right. Ground truth is best viewed in blue lines and the prediction in green. The predicted horizon depth, normal, and gradient are visualized below each panorama, and the ground truth is in the first row. We labeled the significant differences with dashed lines.}
	\label{fig:quligeneral}
\vspace{-0.2cm}
\end{figure*}
\begin{figure*}[htbp]
\centering
\subfigure[Standford 2D-3D \cite{armeni2017joint}]{
\includegraphics[width=0.48\textwidth]{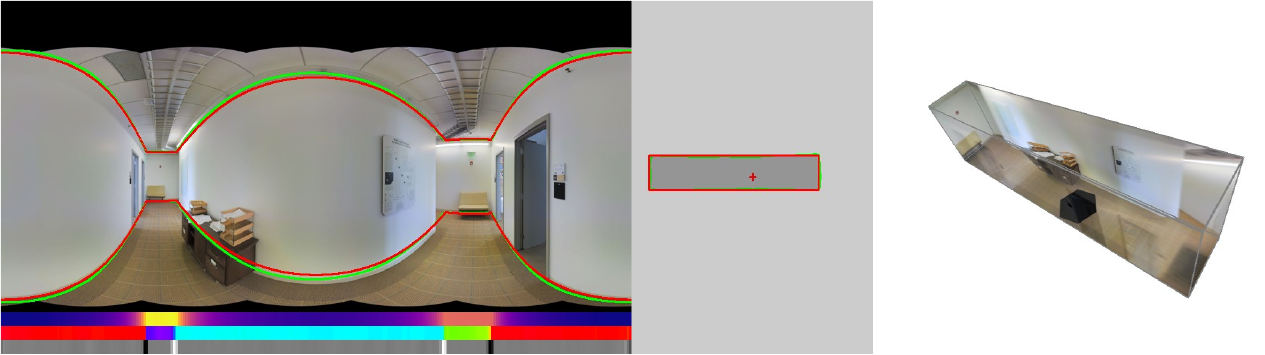}
}%
\subfigure[PanoContext \cite{zhang2014panocontext}]{
\includegraphics[width=0.48\textwidth]{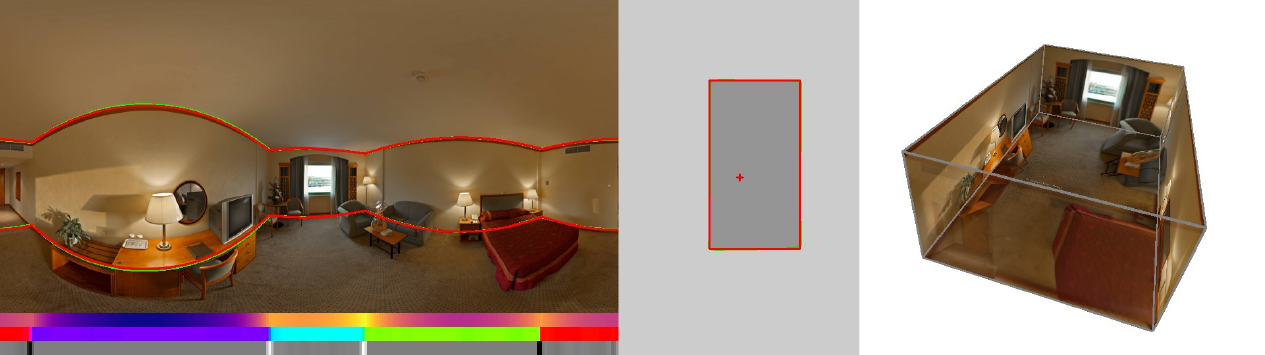}
}%

\subfigure[MatterportLayout \cite{zou2021manhattan}]{
\includegraphics[width=0.48\textwidth]{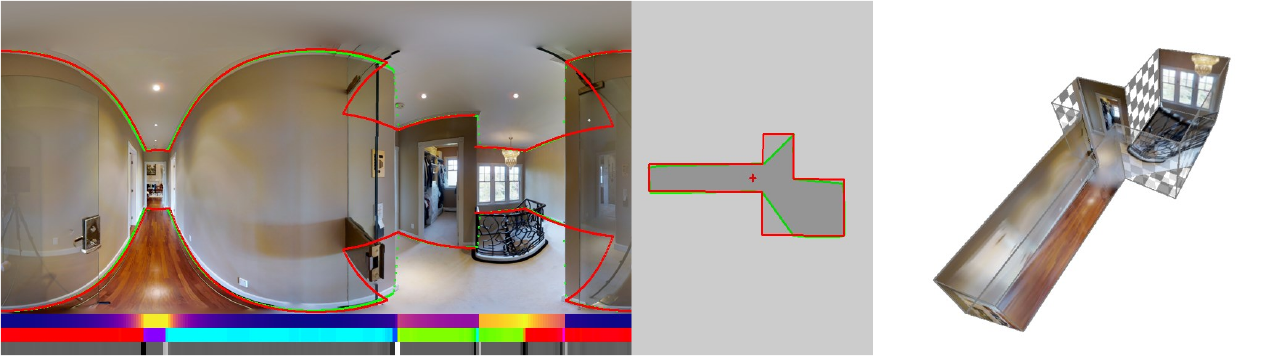}
}%
\subfigure[ZInD \cite{cruz2021zillow}]{
\includegraphics[width=0.48\textwidth]{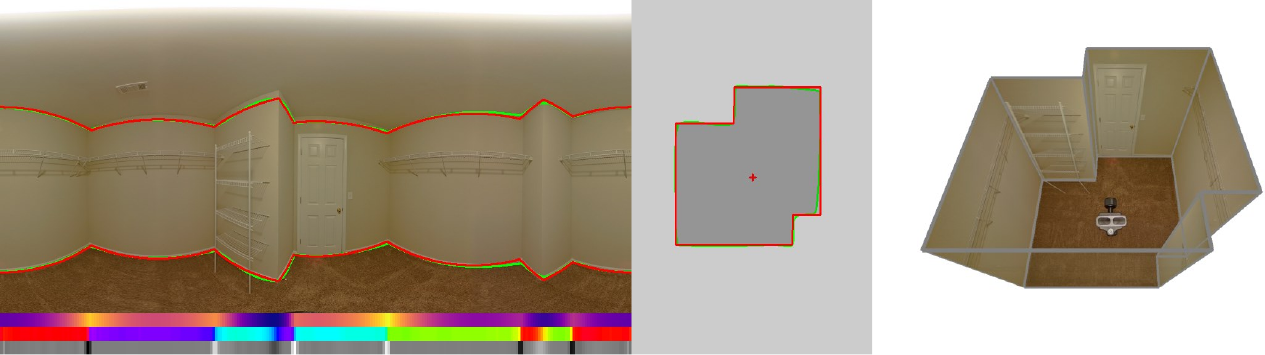}
}%
\centering
\caption{The 3D visualization results. We exhibit the predicted boundaries (best viewed in green lines and the ones with post-processing of the prediction (best viewed in red lines) in the panoramas.}
\label{fig:3dvis}
\vspace{-0.1cm}
\end{figure*}
\begin{figure*}[t]
  \centering
  \includegraphics[width=0.97\textwidth]{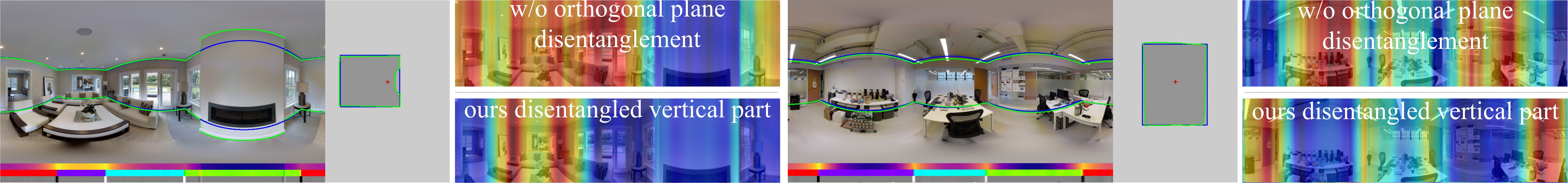} 
  \caption{We exhibit the heat maps of the features of the 1D representation. The one without orthogonal plane disentanglement is shown on the top, and ours is on the bottom.  
  Without disentangling, the top contains redundant and confusing features that are not good for layout estimation. In contrast, our disentangled vertical plane features are more discriminative, showing more attention to the layout corners.
  } 
  \label{fig:visf}
\vspace{-0.2cm}
\end{figure*}
\begin{figure*}[t]
  \centering
  \includegraphics[width=\textwidth]{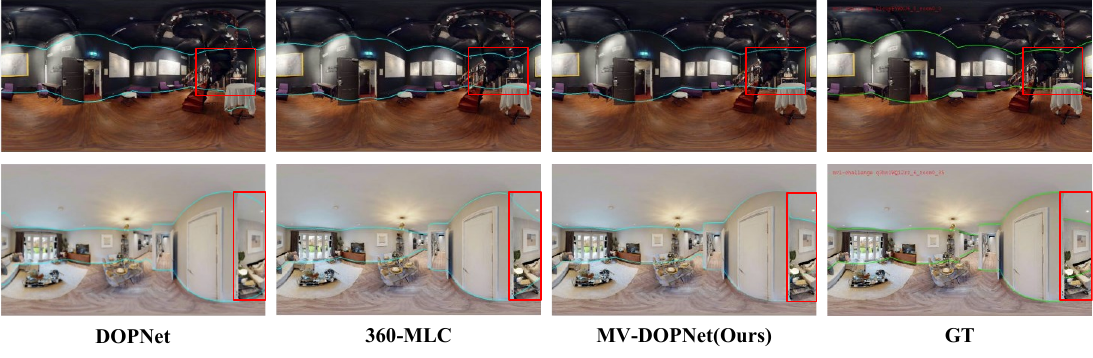} 
  \caption{Qualitative comparison on MVL dataset. We label the significant differences with solid red lines.} 
  \label{fig:occlusion}
\end{figure*}
\subsection{General Room Results}
\label{sec4_4} MatterportLayout \cite{zou2021manhattan} dataset and ZInD \cite{cruz2021zillow} dataset provide more general indoor room layouts, which is much more challenging than the cuboid room layout datasets. Tab. \ref{tab:cp2} exhibits the evaluation on MatterportLayout \cite{zou2021manhattan}/ZInD \cite{cruz2021zillow} datasets. The results of LED$^{2}$-Net \cite{wang2021led} and HorizonNet \cite{sun2019horizonnet} are from \cite{Jiang_2022_CVPR} that we strictly followed. Particularly, Jiang $et$ $al.$ utilize their official code to re-train and re-evaluate with the standard evaluation metrics. 

\noindent\textbf{Comparison Results.} We observe that Dula-Net \cite{yang2019dula} demonstrates much worse performance on the MatterportLayout\cite{zou2021manhattan} dataset, indicating that these perspective view-based methods are hard to adapt to the general indoor scenarios. Moreover, these 1D sequence-based methods are better than those 2D convolution-based schemes. However, they cannot produce accurate 3D layouts from a 1D sequence due to confused plane semantics, even if they introduce more powerful relationship builders (\textit{e.g.}, Bi-LSTM, Transformer). Hence, these methods give worse 3DIoU. Compared with them, our approach offers better performance than all other approaches with respect to 3DIoU because ours captures an explicit 3D geometric cue by disentangling orthogonal planes, which benefits recovering layouts from clear semantics (evidenced by Fig. \ref{fig:visf}).

We show the qualitative comparisons in Fig. \ref{fig:qulicuboid}/\ref{fig:quligeneral}. From the figures, we can observe that our method offers better 3D results (floor plan). In Fig. \ref{fig:3dvis}, our results are very similar to the post-processing version. Both indicate our proposal that disentangling orthogonal planes to capture an explicit 3D geometric cue is an effective strategy.
\begin{table}[t]
\caption{Comparison results with the baseline evaluated on MVL challenge dataset.}
\begin{center}
 \begin{tabular}{l c c c c}
  \toprule
  Method& 2DIoU(\%) &  3DIoU(\%)& RMSE&$\delta_1$\\
    \hline
    \multicolumn{4}{c}{Group 1: pre-trained on MatterportLayout}\\
    \hline
    HorizonNet &78.53&73.93&0.342&0.861\\
    DOPNet&80.71&75.13&0.301&0.839\\
    360-MLC&79.95&76.47&0.350&0.883\\
    360-MLC-Iter&80.87&77.39&0.334&\textbf{0.889}\\
    Ours&\textbf{83.59}&\textbf{79.42}&\textbf{0.268}&0.877\\
     \hline
    \multicolumn{4}{c}{Group 2: pre-trained on ZInd}\\
    \hline
    HorizonNet &63.57&58.37&0.578&0.701\\
    DOPNet&65.35&57.51&0.574&0.740\\
    360-MLC&80.90&76.05&0.329&0.878\\
    Ours&\textbf{83.41}&\textbf{78.82}&\textbf{0.284}&\textbf{0.890}\\
  \bottomrule
 \end{tabular}
\end{center}
 \label{tab:mvl}
 \vspace{-0.2cm}
\end{table}

\subsection{MVL Results}
Diverging from most conventional layout datasets like MatterportLayout \cite{zou2021manhattan} and Zillow \cite{cruz2021zillow}, the dataset employed to evaluate the proposed framework stands out with its expansive rooms, augmented count of rooms within each scene, and a pronounced inclusion of non-Manhattan spaces. These intricacies emphasize a novel spectrum of challenges for room layout estimation, particularly in the absence of labeled data. In this section, we compare the current SoTA methods' performance in facing these dilemmas. All the models are evaluated on the warm-up phase pilot splits. The fine-tuning process is based on the unlabeled challenge phase training splits. Since there is an overlap between the used MVL dataset and the MatterportLayout dataset, we also conduct an additional experiment by applying the parameters pre-trained on the ZInD dataset to demonstrate the multi-view performance more convincingly. 

The comparative results are presented in Table \ref{tab:mvl}. In the table, both "HorizonNet " and "DOPNet" indicate that we just employ the pre-trained parameters without the fine-tuning process. Additionally, the 360-MLC method is enhanced with an iterative optimization strategy (iteratively generating refined pseudo-labels and fine-tuning\footnote{Our solution that won the 1st Place in the MVL Challenge at the CVPR 2023 Workshop.}, denoted as "360-MLC-Iter" in Group 1, Table 3). In the first group, we can observe that our DOPNet performs better than the HorizonNet. This exceptional performance is attributable to two primary factors: first, the incorporation of an advanced horizon-depth representation in our model. Second, the integration of meticulously designed modules within our network. Besides, we observe that our iterative strategy was found to narrow the performance disparities among base layout models. Particularly, models that initially displayed suboptimal performance witnessed notable enhancement. Yet, after the 2nd iteration, a discernible influx of noise disrupts the estimated boundary, preventing it from forming seamless and complete layouts.
However, it is worth noting that even with iterative optimization, the baseline method lagged behind our approach. Our method achieves the top performance, demonstrating the superior effectiveness of the proposed approach.

In Tab.\ref{tab:mvl}, the depth metric (\textit{e.g.}, RMSE, and $\delta_1$) are also reported to demonstrate our multi-view performance. To calculate the depth accuracy metrics, we follow the previous works\cite{Jiang_2022_CVPR}\cite{sun2019horizonnet} to convert the ground truth into horizon-depth representation by setting the camera height as 1.6 meters. Since the horizon-depth representation of layouts is sparse and topological, it is sensitive to the inferior values. Hence, from Group 1 in Table 3, we can observe that while our approach ranks second on $\delta_1$, it achieves a lower RMSE, resulting in more accurate layouts.

\noindent\textbf{Occlusion analysis.} 
From Fig. \ref{fig:occlusion}, we can observe that due to the occlusion (\textit{e.g.}, the stairs, and the walls), the DOPNet fails to produce the correct layouts in these challenging regions. The current state-of-the-art approach that only optimizes the pseudo-labels is also disturbed by the occlusions. In contrast, our approach can benefit from the proposed decision-level optimization and feature-level multi-view perception approach to produce more accurate layouts even in occlusion cases.
\begin{table}[t]
\caption{Ablation study on MatterportLayout \cite{zou2021manhattan} dataset and MVL dataset.}
\begin{center}
 \begin{tabular}{l c c}
  \toprule
  Method& 2DIoU(\%) &  3DIoU(\%)\\
    \hline
    \multicolumn{3}{c}{Group 1: ablation on MatterportLayout (Monocular)}\\
    \hline
    w/o Cross-scale interaction&83.24&80.71\\
    w/o Feature assembling&82.36&80.10\\
    \hline
    w/o Disentangling planes&83.23&80.60\\
    w/o Flipping fusion&83.01&80.35\\
    \hline
    w/o Discriminative channels&82.74&80.23\\
    w/o Long-range dependencies&83.04&80.97\\
    w/o Residuals&82.89&80.45\\
    \hline
    Ours [Full]&\textbf{83.46}&\textbf{81.34}\\
    \hline
    \multicolumn{3}{c}{Group 2: ablation on MVL (Multi-view)}\\
    \hline
    DOPNet[mp3d]&80.71&75.13\\
    DOPNet[fine-tuning]&81.64&51.91\\
    MV-DOPNet[fine-tuning]&82.15&53.99\\
    DOPNet[w/ $\mathcal{L}_{r}$]&81.70&76.02\\
    MV-DOPNet[w/ $\mathcal{L}_{r}$]&82.44&77.72\\
    MV-DOPNet[w/ $\mathcal{L}_{r}, \mathcal{L}_{n}^w, \mathcal{L}_{g}^w$]&\textbf{83.59}&\textbf{79.42}\\
  \bottomrule
 \end{tabular}
\end{center}
 \label{tab:abla}
 \vspace{-0.2cm}
\end{table}
\subsection{Ablation study}
\label{section42}
 We exhibit ablation studies in Tab. \ref{tab:abla} group 1, where each component of our model is evaluated on MatterportLayout \cite{zou2021manhattan}. To demonstrate the effectiveness of the proposed feature assembling mechanism, we first ablate cross-scale interaction in the distortion-aware feature aggregation module (denoted as "w/o Cross-scale interaction"). Then, we further remove the distortion elimination (denoted as "w/o Feature assembling"). 
To exhibit the benefit of orthogonal plane disentanglement, we remove the proposed disentangling orthogonal planes procedure (denoted as "w/o Disentangling planes") and soft-flipping fusion, respectively. 
Finally, we show the effect of each attention mechanism (denoted as "w/o Discriminative channels", "w/o Long-range dependencies", and "w/o residuals", respectively). For a fair comparison, we train each model with the same epochs (500).

\noindent\textbf{Cross-scale Distortion Awareness.} 
 From Tab. ~\ref{tab:abla}, the pipeline without cross-scale interaction shows inferior performance to the complete model. Further removing the distortion elimination part, the pipeline yields worse results. It is proved that both dealing with distortions and integrating cross-scale features are essential for layout estimation. 

\noindent\textbf{Pre-segmenting Orthogonal Planes.} Since the geometric cues are essential for inferring 3D information from 2D images, we propose to disentangle orthogonal planes. To prove its effectiveness, we remove that stage but preserve the flipping fusion strategy (the effectiveness of this strategy is validated separately). The results in Tab. ~\ref{tab:abla} show that this stage can make the performance more competitive. Besides, the embedding of symmetry property also contributes to the promotion of the predicted results.

\noindent\textbf{Reconstructing 1D Representations.} 
The dependencies of the 1D sequences have changed when disentangling the orthogonal planes. Hence, we propose to reconstruct the 1D representations. We verify the effectiveness of each attention mechanism in turn. From Tab. \ref{tab:abla}, we can observe that all three reconstruction operations (generating discriminative channels, rebuilding long-range dependencies, and providing the missing residuals) can benefit the overall performance. Intuitively, when removing the channel-wise graph, the pipeline's performance decreases significantly, demonstrating that enforcing the network concentration on discriminative channel information can capture effective information by avoiding redundancy.

As shown in Fig. \ref{fig:visf}, we can observe that the 1D representation of the layout produced by our DOPNet is much more precise than that produced by the baseline method.

\noindent\textbf{MVL Specific Designs.} From Tab. \ref{tab:abla} group 2, we can observe that multi-view insights elevate our model's 2DIoU score to the forefront, while our formulated ratio constraints simultaneously enhance both 3DIoU and 2DIoU scores. Additionally, the improved $\mathcal{L}_{n}^w$,$ \mathcal{L}_{g}^W$ losses further boost overall model performance, underscoring our approach's effectiveness in advancing state-of-the-art technologies. 
\section{Limitation}
Although the proposed MV-DOPNet can achieve effective multi-view perception, our approach still has some limitations. One limitation is that the advanced horizon-depth/ratio representation adopted in our methodology is based on an important assumption (\textit{i.e.}, a normalized and fixed camera height, 1.6 meters) which we must follow to recover the layouts. Besides, our approach follows the 360-MLC\cite{solarte2022mlc} method, which still relies on a pre-trained model. Since the room layout boundary representation is inherently sparse and topological, when we attempted to leverage multi-view projection consistency constraints for completely unsupervised learning, the model failed to converge and could not recover the room layout from the generated layout representation. In the future, we will try to solve this problem by proposing more suitable layout representations (\textit{e.g.}, an equivalent but dense representation).
\section{Conclusion}
\label{section5}
In this paper, we propose a framework, for multi-view layout estimation. Current approaches generate a 1D representation by a vertical compression operation. We argue that this strategy confuses the semantics of different planes. We propose to disentangle orthogonal planes to get spotless 1D representations, as well as capture geometric cues in 3D space. Specifically, we obtain disentangled semantics by eliminating distortions, pre-segmenting orthogonal plans, and reconstructing dependencies of sequences. Besides, we propose both a horizon-depth optimizer and a multi-view specific module to fully exploit the multi-view layout consistency both at the decision level and the feature level. Our experimental results demonstrate that dealing with distortion, as well as integrating shallow and deep features, can enhance the performance. Experiments demonstrate that our algorithm significantly outperforms current SoTA methods on both monocular layout estimation and multi-view layout estimation tasks.
\ifCLASSOPTIONcompsoc
  \section*{Acknowledgments}
\else
  \section*{Acknowledgment}
\fi

We would like to thank Wei Wang, Shujun Huang, Yue Zhan, and Shuai Zheng for valuable discussions. Thanks to all authors for their insightful research and open source projects. This work was supported by the National Natural Science Foundation of China (Nos. 62172032,  62120106009).

\ifCLASSOPTIONcaptionsoff
  \newpage
\fi



%

{\small
\normalem
\bibliographystyle{IEEEtran}
\bibliography{egbib}
}

%



\end{document}